\documentclass[journal]{IEEEtran}

\usepackage{booktabs}
\usepackage{xcolor}
\usepackage{colortbl}
\usepackage{amssymb}
\usepackage{algorithmic}
\usepackage{amsmath} 
\usepackage{amsfonts}
\usepackage[linesnumbered,ruled,lined]{algorithm2e}
\ifCLASSINFOpdf
\usepackage[pdftex]{graphicx}
\else
\usepackage[dvips]{graphicx}
\fi
\usepackage{epstopdf}
\usepackage{array}
\ifCLASSOPTIONcompsoc
\usepackage[caption=false,font=normalsize,labelfont=sf,textfont=sf]{subfig}
\else
\usepackage[caption=false,font=footnotesize]{subfig}
\fi
\usepackage{float} 
\usepackage{fixltx2e}
\usepackage{color}
\usepackage{xcolor}
\usepackage{textcomp}
\usepackage[normalem]{ulem}

\usepackage{cite}
\usepackage{multibib} % necessary for BSTcontrol
\usepackage{nomencl}
\makenomenclature
\usepackage{etoolbox}
\renewcommand\nomgroup[1]{%
	\item[\bfseries
	\ifstrequal{#1}{A}{Indices and Sets}{%
		\ifstrequal{#1}{B}{Continuous Decision Variables}{%
			\ifstrequal{#1}{C}{Discrete Decision Variables}{       \ifstrequal{#1}{D}{Parameters}{
			        \ifstrequal{#1}{E}{Dynamic System Variables and Parameters}} }}} ]}
\usepackage{hyperref}
\hypersetup{
	colorlinks=true,
	linkcolor=blue,
	filecolor=magenta,      
	urlcolor=cyan,}
\makeatletter
\newcommand\footnoteref[1]{\protected@xdef\@thefnmark{\ref{#1}}\@footnotemark}
\makeatother
\usepackage{listings}

\IEEEoverridecommandlockouts

\begin{document}
\bstctlcite{BSTcontrol}%externally control some aspects of the bibliography style (use multibib)
	
\title{Deep Active Learning for Solvability Prediction in Power Systems}

\author{Yichen~Zhang,~\IEEEmembership{Member,~IEEE,}
	~Jianzhe~Liu,~\IEEEmembership{Member,~IEEE,}
	~Feng~Qiu,~\IEEEmembership{Senior~Member,~IEEE,}
	~Tianqi~Hong,~\IEEEmembership{Member,~IEEE,}
	~Rui~Yao,~\IEEEmembership{Senior~Member,~IEEE}
	\thanks{
	This work was supported by the U.S. Department of Energy Office of Electricity -- Advanced Grid Modeling Program.
	
	Y. Zhang, J. Liu, F. Qiu, T. Hong and R. Yao are with Argonne National Laboratory, Lemont, IL 60439 USA. Email:yichen.zhang@anl.gov).
}}

%\markboth{IEEE TRANSACTIONS ON POWER SYSTEMS,~Vol.~-, No.~-, -~20--}%
%{Shell \MakeLowercase{\textit{et al.}}: Bare Demo of IEEEtran.cls for IEEE Journals}
\maketitle

\begin{abstract}
Traditional methods for solvability region analysis can only have inner approximations with inconclusive conservatism and handle limited types of power flow models. In this paper, we propose a deep active learning framework for power system solvability prediction. Compared with the passive learning methods where the training is performed after all instances are labeled, active learning selects most informative instances to be label and therefore significantly reduces the size of the labeled dataset for training. In the active learning framework, the acquisition functions, which correspond to different sampling strategies, are defined in terms of the on-the-fly posterior probability from the classifier. The IEEE 39-bus system is employed to validate the proposed framework, where a two-dimensional case is illustrated to visualize the effectiveness of the sampling method followed by the full-dimensional numerical experiments.
\end{abstract}


\section{Introduction}
%Power system instability can be mainly categorized into static and dynamic instability. On the one hand, static instability occurs when the system state exceeds the solvability boundary, leading to no solution of the power flow. This process is closely related to voltage collapse. On the other hand, dynamic instability happens when the system trajectory goes belong the stability boundary, also known as the region of attraction. This evolution is closely related to rotor-angle stability. Exactly characterizing the region of solvability in the parameter space and the region of attraction in the state space are both open problems in power systems.

%Machine learning techniques have long been employed to tackle these challenges, where sampling over appropriate subspaces is an essential prerequisite to generate high-quality training data. For dynamic stability analysis, since the region of attraction is defined around an stable equilibrium point, sampling can be effectively initialized using the equilibrium point information. Unfortunately, it is not the case for solvability analysis. 

Power system under the stochastic power injections of renewable energy may exceed the loadability limits and result in voltage collapse. Therefore, it is important to quickly assess if power flow has a solution (i.e. solvable) given a set of power injections. 
The conventional approach is to solve the power flow equations
numerically using iterative methods. While many real-time operation scenarios desire non-iterative and analytical approaches to determine the solvability. Earlier research focused on solvability condition of decoupled power flow models \cite{wu1982steady}\cite{ilic1992network}. The fixed-point theorem has been used to obtain the solvability of coupled full power flow models in distribution networks \cite{bolognani2015existence}.
Improvements from \cite{bolognani2015existence} have been achieved in \cite{wang2016explicit,nguyen2018framework,dvijotham2017solvability}. Ref. \cite{cui2019solvability} derived a seminal explicit sufficient solvability condition that certify existence and uniqueness of solutions, which dominates earlier works \cite{bolognani2015existence,wang2016explicit,nguyen2018framework,dvijotham2017solvability}.

Despite these innovative works, state-of-the-art analytical condition still cannot handle coupled full power flow models with different types of buses. The most recent work in \cite{cui2019solvability} can handle a system with only slack and PQ buses\footnote{A bus that the active and reactive power injections are fixed}. While, PV buses\footnote{A bus that the active power injection and voltages are fixed} are common in power flow analysis, but cannot be considered in most analytical conditions. 

Machine learning techniques have long been employed to amend the shortcomings of analytical methods. The recent success of deep learning has facilitated its application into power flow problems \cite{Du2019, Pan2020, Hu2020, Lei2020}. The N-1 contingency screening using a deep convolutional neural network is presented in \cite{Du2019}. Since it focused on the contingency screening, the load power injections are fixed. The security-constrained DC optimal power flow is solved under the aid of deep learning in \cite{Pan2020}. Refs. \cite{Hu2020} and \cite{Lei2020} propose physics-informed learning models to solve the power flow. The results are promising, but the model's capability under the full power injection space is not demonstrated. In a nutshell, existing works cannot be generalized into the solvability problem since the \emph{AC power flow under the full power injection space has not been investigated}.

To this end, we propose a deep learning-based approach to predict the solvability. Our method consists of two phases: off-line training and online prediction. In the off-line training phase, we sample power injections over all permissible ranges. This results in very high volumes of samples. Simultaneously, the labeling process requires solving the AC power flow problem of all samples and demands considerable computation resources. Therefore, we employ the active learning framework|a family of machine learning methods which \emph{query} the data instances to be labeled for training by an \emph{oracle} (e.g., a human annotator)|to achieve higher accuracy with much fewer labeled examples than passive learning for solvability prediction. Active learning integrates intelligent sampling and machine learning as a closed loop, and it is valuable in problems where unlabeled data are available but obtaining training labels is expensive. Although sampling towards more informative subspaces has been studied in \cite{krishnan2013importance}, closed-loop integration of machine learning and intelligent sampling like active learning has not been explored yet. This is the first paper to use deep active learning for solvability prediction to the authors' best knowledge.

%Deep neural networks (DNNs) achieved unprecedented success for many supervised learning tasks such as image classification and object detection. Although DNNs are successful in many scenarios, there still exists an obvious limitation: the requirement for a large set of labeled data. To address this issue, Active Learning (AL) appears as a compelling solution by searching the most informative data points (batch) to label from a pool of unlabeled samples in order to maximize prediction performance.

\section{Deep Active Learning for Power Flow Solvability Approximation}
% The AC power flow equations are described as follows. Real and reactive power balances at each bus $i\in\mathcal{B}$ read
% \begin{align}
% \label{eq_power_balance}
% \begin{aligned}
% &\sum_{g\in\mathcal{G}_{i}}p_{g}^{\text{G}} = \sum_{d\in\mathcal{D}_{i}}p_{d}^{\text{D}} + \sum_{l\in\mathcal{L},j\in\mathcal{B}_{j}:j\neq i}p_{l}^{ij}+G_{i}^{\text{B}}v_{i}^{2}\\
% &\sum_{g\in\mathcal{G}_{i}}q_{g}^{\text{G}} = \sum_{d\in\mathcal{D}_{i}}q_{d}^{\text{D}} + \sum_{l\in\mathcal{L},j\in\mathcal{B}_{j}:j\neq i}q_{l}^{ij}-B_{i}^{\text{B}}v_{i}^{2}
% \end{aligned}
% \end{align}
% A line $l\in\mathcal{L}$ connects bus $i\in\mathcal{B}_{l}$ to  $j\in\mathcal{B}_{l}$, where $j\neq i$. The power flows from bus $i$ to $j$ are
% \begin{align}
% \label{eq_power_flow}
% \begin{aligned}
% & p_{l}^{ij}=g_{l}^{ij}v_{i}^{2} - g_{l}^{ij}v_{i}v_{j}\cos\theta_{ij} - b_{l}^{ij}v_{i}v_{j}\sin\theta_{ij}\\
% & q_{l}^{ij}=(-b_{l}^{ij} - 0.5b_{l}^{C})v_{i}^{2}  + b_{l}^{ij}v_{i}v_{j}\cos\theta_{ij} - g_{l}^{ij}v_{i}v{j}\sin\theta_{ij}
% \end{aligned}
% \end{align}
% In Eqs. (\ref{eq_power_balance}) and (\ref{eq_power_flow}), the active and reactive power injections are input variables; the voltages are system state variables; the power flow are intermediate variables.
Consider an $N_{\text{B}}$-bus power network with $N_{\text{G}}$ generator buses and $N_{\text{D}}$ load buses. Let $\mathcal{N}_{\text{B}}$, $\mathcal{N}_{\text{G}}$ and $\mathcal{N}_{\text{D}}$ denote the set of all buses, generator buses and load buses, respectively. $\mathcal{N}_{PQ}$ and $\mathcal{N}_{PV}$ represent sets of PQ and PV buses, respectively. The AC power flow equations are as follows
\begin{align}
\label{eq_power_flow}
\begin{aligned}
&\resizebox{1\hsize}{!}{$\displaystyle{P_{i}^{\text{G}}-P_{i}^{\text{D}}=\sum_{j\in\mathcal{N}_{\text{B}}}V_{i}V_{j}(G_{ij}\cos\theta_{ij} + B_{ij}\sin\theta_{ij}), i\in\mathcal{N}_{PQ}\cup\mathcal{N}_{PV} }$}\\
&Q_{i}^{\text{G}}-Q_{i}^{\text{D}}=\sum_{j\in\mathcal{N}_{\text{B}}}V_{i}V_{j}(G_{ij}\sin\theta_{ij} - B_{ij}\cos\theta_{ij}), i\in\mathcal{N}_{PQ}
\end{aligned}
\end{align}
The bus voltages $V_{i}$ and angles $\theta_{i}$ are system state variables, and $\theta_{ij}=\theta_{i}-\theta_{j}$. The existence of solutions to Eq. (\ref{eq_power_flow}) depends on the values of power injections. To this end, the goal is to build a classifier using a multi-layer perception (MLP) model that can separate the solvable power injections from the non-solvable ones. Therefore, the inputs to the deep neural network are power injections defined as follows
\begin{align*}
\begin{aligned}
&\bold{X} = \\
&\left[\begin{array}{c}
\bold{P}_{1}^{\text{G}},\cdots,\bold{P}^{\text{G}}_{N_{\text{G}}}, % Note I have a comment here
\bold{Q}_{1}^{\text{G}},\cdots,\bold{Q}^{\text{G}}_{N_{\text{G}}},
\bold{P}_{1}^{\text{D}},\cdots,\bold{P}^{\text{D}}_{N_{\text{D}}}
\bold{Q}_{1}^{\text{G}},\cdots,\bold{Q}^{\text{G}}_{N_{\text{D}}}
\end{array}\right] \label{eq_dnn_input}
\end{aligned}
\end{align*}
where $\bold{P}_{g}^{\text{G}}$ and $\bold{Q}_{g}^{\text{G}}$ are samples of generation active and reactive power injections at generator bus $g$; $\bold{P}_{d}^{\text{D}}$ and $\bold{Q}_{d}^{\text{D}}$ are samples of load active and reactive power injections at load bus $d$. Let $s$, $N_{\text{S}}$ and $\mathcal{N}_{\text{S}}$ denote sample index, sample numbers and the set of sample indices, respectively. Each sample in $\bold{X}$, denoted as $\bold{x}_{s}=\bold{X}_{[s,:]}$\footnote{The subscript $[s,:]$ denotes the $s$th row of a matrix.}, will be solved by PSS/E and label its solvability: 0 indicates $\bold{X}_{[s,:]}$ is not solvable and 1 otherwise. To ensure that the generator reactive power samples remain the same in the final solution, we set the reactive power upper and lower limits of all generators to be equal to corresponding samples. We denote the classes of non-solvable and solvable as $C_{q}$ for $q=1,2$, respectively. The label data after one-hot encoding reads
\begin{align}
\bold{Y}^{*} =
\left[\begin{array}{c}
\bold{y}^{*}_{1}\\\vdots\\\bold{y}^{*}_{N_{\text{S}}}
\end{array}\right],
\bold{y}^{*}_{s} =
\left[\begin{array}{cc}
y^{*}_{s,1},y^{*}_{s,2}
\end{array}\right]
\end{align}
where $y^{*}_{s,q}=1$ indicates that sample $s$ belongs to class $q$. We then apply probabilistic smoothing approximations to the discrete label values. It is well known that when the targets are one-hot encoded and an appropriate loss function is used, an MLP directly estimates the posterior probability of class membership $C_{q}$ conditioned on the input variables $\bold{x}_{s}$, denoted by $p(C_{q}|\bold{x}_{s})$. Denote the MLP classifier as $\bold{y}_{s} = f(\bold{x}_{s};\theta)$ where $\bold{y}_{s}=[y_{s,1},y_{s,2}]=[p_{\theta}(C_{1}|\bold{x}_{s}),p_{\theta}(C_{2}|\bold{x}_{s})]$, where $p_{\theta}(C_{q}|\bold{x})$ for $q=1,2$ denotes the posterior probability of class membership $q$ given by the classifier under parameter $\theta$. The network parameters $\theta$ can be calculated using the maximum likelihood estimation. Therefore, we minimize the negative logarithm of the likelihood function, known as the cross-entropy loss, as follows
\begin{align}
L=-\frac{1}{N_{\text{s}}}\sum_{s=1}^{N_{\text{s}}}\sum_{q=1}^{2}y^{*}_{s,q}\log y_{s,q}
\end{align}
Since the output values of the MLP are interpreted as probabilities, they each must lie in the range (0,1), and they must sum to unity. This can be achieved by using a \emph{softmax} activation function at the output layer of the MLP.

\section{Active Learning Framework}
Assume we randomly generate a feature set $\mathcal{X}$ that is sufficiently large to represent the underlying physical features. In traditional \emph{passive} supervised learning methods, we will generate labels for the entire feature set $\mathcal{X}$, denoted as $\mathcal{Y}_{\mathcal{X}}$, using the simulation software and result parser, which is regarded as the oracle. The labeling process is computationally demanding if the data set is large and becomes intractable for high-dimensional problems. This is known as the \emph{labeling bottleneck}, which occurs not only in power systems but also in computer vision, natural language processing, and other machine learning tasks. The active learning framework can overcome such a labeling bottleneck. A typical active learning algorithm consists of the following steps
\begin{enumerate}
	\item Define a supervised learning model $f(\textbf{x};\theta)$
	\item Randomly select a small set $\mathcal{L}\in\mathcal{X}$ and generate the labels $\mathcal{Y}_{\mathcal{L}}$ accordingly
	\item Train the model $f$ using the feature-label pairs $\textless\mathcal{L}, \mathcal{Y}_{\mathcal{L}}\textgreater$
	\item Query instances $\mathcal{Q}$ from a large unlabeled pool $\mathcal{U}$, where $\mathcal{U}=\mathcal{X}\setminus\mathcal{L}$, generate the labels $\mathcal{Y}_{\mathcal{Q}}$ accordingly, and move $\mathcal{Q}$ from $\mathcal{U}$ to $\mathcal{L}$ 
	\item Train the model $f(\textbf{x};\theta)$ using the updated feature-label pairs $\textless\mathcal{L}, \mathcal{Y}_{\mathcal{L}}\textgreater$
	\item Repeat Steps (4)-(5) until the termination criterion is met
\end{enumerate}
The algorithm's pseudocode is formally presented in Algorithm \ref{algo_bmal}. Obviously, the querying strategy $a(\cdot,\cdot)$ differentiates the active learning from passive learning algorithms. In other words, active learning under the random querying strategy will be equivalent to passive learning algorithms. The queries can be either selected in serial (one at a time) or batches (several to be labeled at once). Algorithm \ref{algo_bmal} is the batch-mode active learning. Given the machine learning model $f$, unlabeled pool $\mathcal{U}$, and inputs $\textbf{X}\in\mathcal{U}$, the querying strategy can be represented as a function $a$, which is referred to as the \emph{acquisition function} written as follows
\begin{align}
\textbf{x}^{*}=\arg\max_{\textbf{X}\in\mathcal{U}}a\left( \textbf{X}, f(\cdot;\theta)\right) 
\end{align}
where $\textbf{x}^{*}$ denotes the most informative sample selected by the corresponding strategy. 

\begin{algorithm*}
	\SetKwData{Left}{left}\SetKwData{This}{this}\SetKwData{Up}{up}
	\SetKwFunction{Union}{Union}\SetKwFunction{Oracle}{Oracle}\SetKwFunction{Train}{Train}
	\SetKwInOut{Input}{input}\SetKwInOut{Output}{output}
	
	\Input{Labeled set $\mathcal{L}$, unlabeled set $\mathcal{U}$, query strategy $a(\cdot,\cdot)$, query batch size $B$, labeling oracle $\Oracle(\cdot)$, deep neural net $f(\cdot;\theta)$, neural network training function $\Train(\cdot,\cdot)$}
	
	%	\BlankLine
	$\mathcal{A} \leftarrow \emptyset$ \tcp*[f]{initialize the set to store acquisition instances}
	
	\Repeat{some stopping criterion}{
		
		$\theta \leftarrow \Train(\mathcal{L},f(\cdot;\theta))$ \tcp*[f]{train the model using current $\mathcal{L}$}
		
		\For{$i \leftarrow 1$ \KwTo $B$}{
			
			$\textbf{x}^{*}_{i} \leftarrow \arg\max_{\mathcal{U}}a(\mathcal{U}, f(\cdot;\theta))$ \tcp*[f]{query the instance from the unlabeled set}
			
			$\textbf{y}^{*}_{i} \leftarrow \Oracle(\textbf{x}^{*}_{i})$  \tcp*[f]{label the acquisition instance}
			
			$\mathcal{L} \leftarrow \mathcal{L}\cup\textless\textbf{x}^{*}_{i},\textbf{y}^{*}_{i}\textgreater$ \tcp*[f]{add the labeled query to $\mathcal{L}$}
			
			$\mathcal{U} \leftarrow \mathcal{U}-\textbf{x}^{*}_{i}$ \tcp*[f]{remove the labeled query from $\mathcal{U}$}
			
			$\mathcal{A} \leftarrow \mathcal{A}\cup\textless\textbf{x}^{*}_{i},\textbf{y}^{*}_{i}\textgreater$ \tcp*[f]{store the acquisition instance}
		}
	}
	\Output{Trained deep neural net $f(\cdot;\theta)$, all acquisition instances $\mathcal{A}$}
	\caption{Batch-mode active learning algorithm}\label{algo_bmal}
\end{algorithm*}

\subsection{Query Strategy: Uncertainty Sampling}
The query strategy aims at evaluating the informativeness of unlabeled instances. There have been many proposed ways of formulating such query strategies in the literature. Interested readers can refer to \cite{settles2009active} for more details. Among all frameworks, the most widely used and computationally efficient methodology is uncertainty sampling. In this letter, an active learner queries the most difficult instances to classify by the deep learning model trained at the current stage. When interpreting the binary classification using a probabilistic model, uncertainty sampling queries the instance whose posterior probability provided by the classifier is the closest to 0.5 \cite{settles2009active}. In other words, the selected sample is the least confident to the classifier. For a general multi-class problem, this \emph{least confident sampling} \cite{settles2009active} can be expressed as follows
\begin{align}
\label{eq_least_confident}
\begin{aligned}
\textbf{x}^{*}_{\text{LC}}=\underset{\textbf{x}\in\mathcal{U}}{\mathrm{argmin}}\max_{q=1,2}p_{\theta}(C_{q}|\bold{x})
\end{aligned}
\end{align}
In the case of multi-class classification, this metric omits information about the remaining labels. To compensate this omission, the \emph{margin sampling} is introduced as follows 
\begin{align}
\label{eq_margin_sampling}
\begin{aligned}
\textbf{x}^{*}_{\text{M}}=\underset{\textbf{x}\in\mathcal{U}}{\mathrm{argmin}}| p_{\theta}(C_{2}|\bold{x})-p_{\theta}(C_{1}|\bold{x})|
\end{aligned}
\end{align}
Besides the aforementioned metrics, the \emph{entropy sampling} is also widely used to measure the amount of information that is encoded and can be only as a metric in active learning
\begin{align}
\label{eq_entropy_sampling}
\begin{aligned}
\textbf{x}^{*}_{\text{E}}=\underset{x}{\mathrm{argmax}}\left( -\sum_{q=1}^{2}p_{\theta}(C_{q}|\bold{x})\log p_{\theta}(C_{q}|\bold{x})\right) 
\end{aligned}
\end{align}
As pointed out in \cite{settles2009active} and many other references, although all strategies generally outperform passive baselines, the best strategy may be application-dependent. Thus, we apply all three strategies to the solvability problem in this letter.

\section{Case Study}\label{sec_case}
\begin{figure}[h]
	\centering
	\includegraphics[scale=0.7]{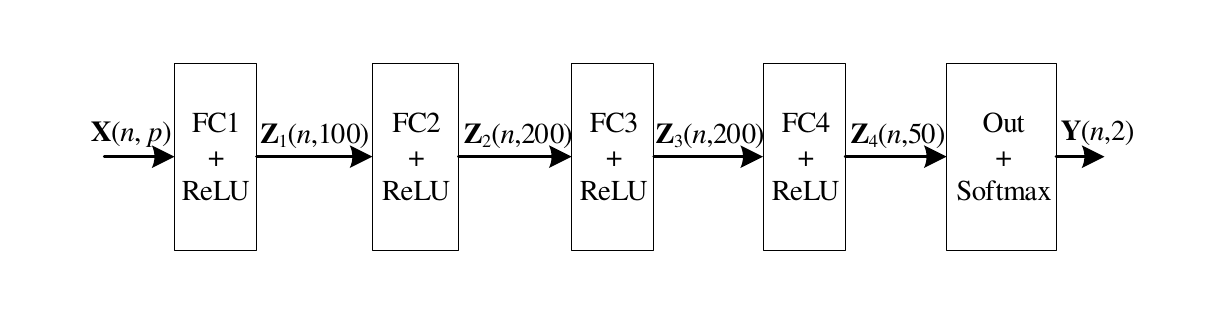}
	\caption{The structure of the deep neural network.}
	\label{fig_dnn}
\end{figure}

We use the IEEE 39-bus system to demonstrate the approach. The deep neural net is shown in Fig. \ref{fig_dnn}. The PSS/E software and the Newton method is used to label the sample. Theoretically speaking, the certificate from the PSS/E software is not sufficient and necessary conditions of solvability. However, considering the fact that the sufficient and necessary conditions for full model power flow solvability with mixed PV and PQ buses are still open problems, we believe that labels from the most widely-used tool in the power community could provide sufficient trustworthy results to guide system operators.

During the training, we also face the data unbalance scenarios as the number of unsolvable samples is larger than the one of solvable samples. Classification accuracy, which is the most-used metric for evaluating classification models, can be misguiding under this circumstance, as high metrics cannot guarantee prediction capacity for the minority class. Here, we employ the under-sampling strategy to resolve this issue. With under-sampling, we randomly remove a subset of samples from the class with more instances to match the number of samples coming from each class. In the active learning algorithm, the under-sampling step takes place after the \texttt{Oracle} labels all selected samples.

\subsection{Solvability Region of a Two-dimensional Case}
\begin{figure}[h]
    \centering
    \includegraphics[scale=0.3]{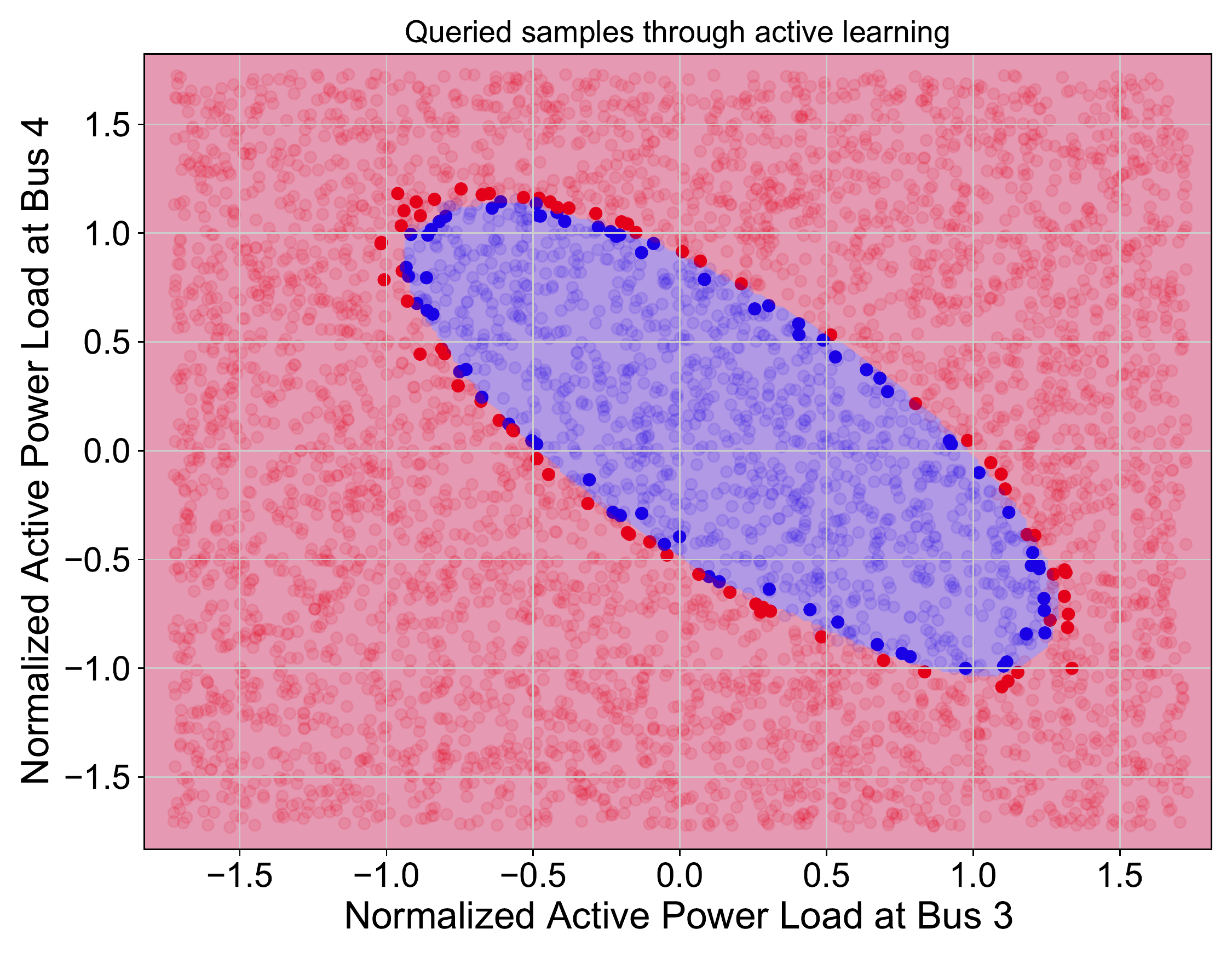}
    \caption{Queried instances by the margin sampling strategy, which precisely selects instances at the solvability boundary. The blue area denotes the solvability region predicted by the neural network. Labeled dataset in the background indicates the estimation is not conservatism.}
    \label{fig_2d}
\end{figure}

First, we illustrate a two-dimensional case for visualization purposes. In this case, we uniformly sample active power loads at Buses 3 and 4 from $-3000$ MW to $3000$ MW. Before the training starts, all samples are normalized. We allocate 80\% samples for training and 20\% samples for testing. The active learner randomly selects 100 samples from the training dataset to label for the initial training phase and queries ten instances in each iteration using the margin sampling strategy. The algorithm terminates if the averaged testing accuracy of the last four iterations is greater than 95\% or the algorithm reaches 30 iterations. The margin sampling strategy terminates after seven iterations, and achieves 95.3\% accuracy with only 170 labeled samples. While the random strategy fails to meet the accuracy criterion after 30 iterations, achieving only 94.6\% accuracy with 400 labeled samples. Samples that are queried by the active learner are plotted in Fig. \ref{fig_2d}, where the decision boundary of the neural network is illustrated using the colored areas (the blue area is solvable). Meanwhile, labeled dataset in the background indicates the estimation is not conservatism. As one can observe, the margin sampling strategy precisely selects instances at the solvability boundary, indicating significantly high sampling efficiency. 

\subsection{Solvability Region under Full Power Injections}
\begin{figure}[h]
    \centering
	\includegraphics[scale=0.24]{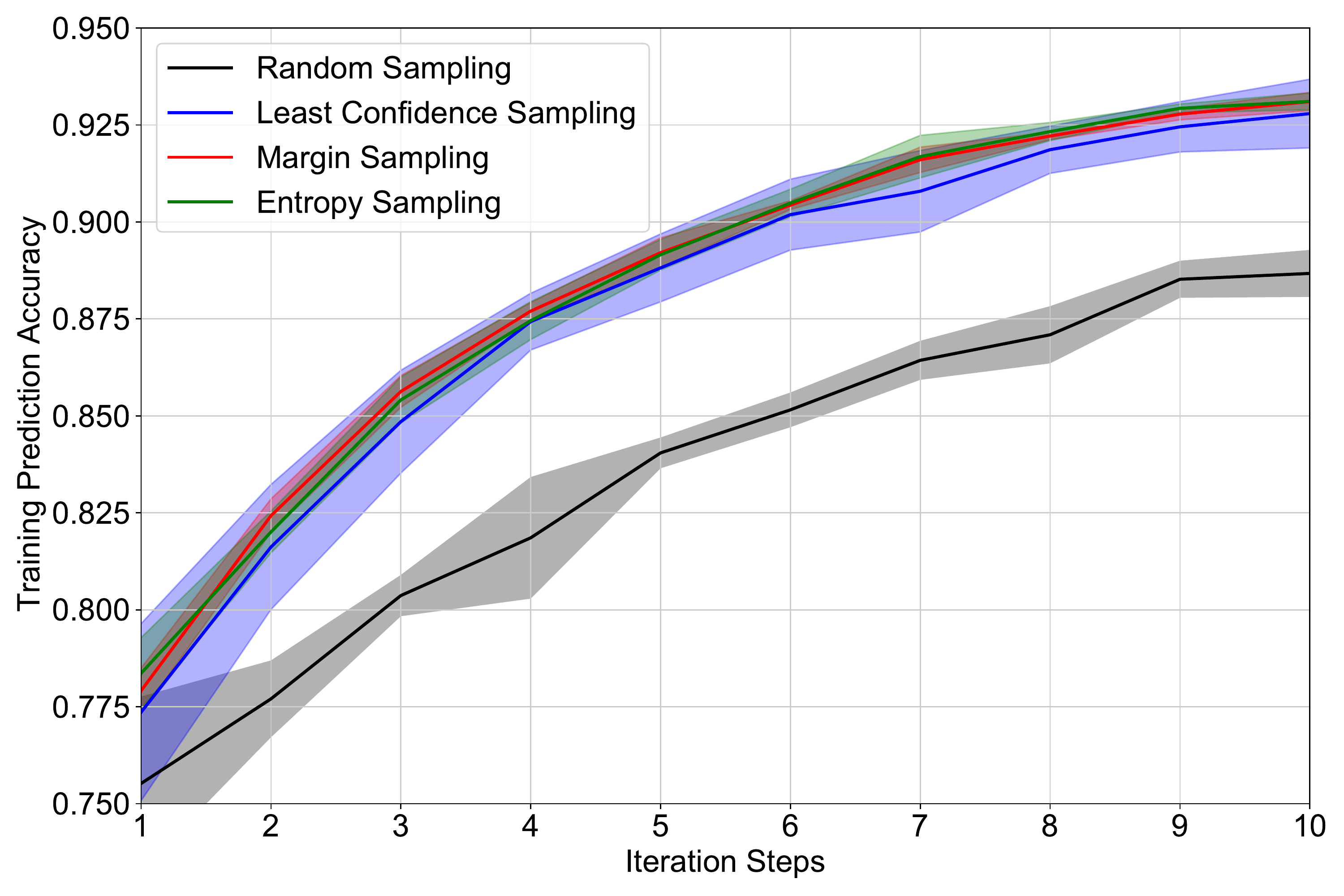}
	\caption{Training accuracy of different sampling strategies. The solid lines indicate the mean values while the shaded areas represent the standard deviations.}
	\label{fig_al}
\end{figure}

\begin{figure}[h]
	\centering
	\includegraphics[scale=0.24]{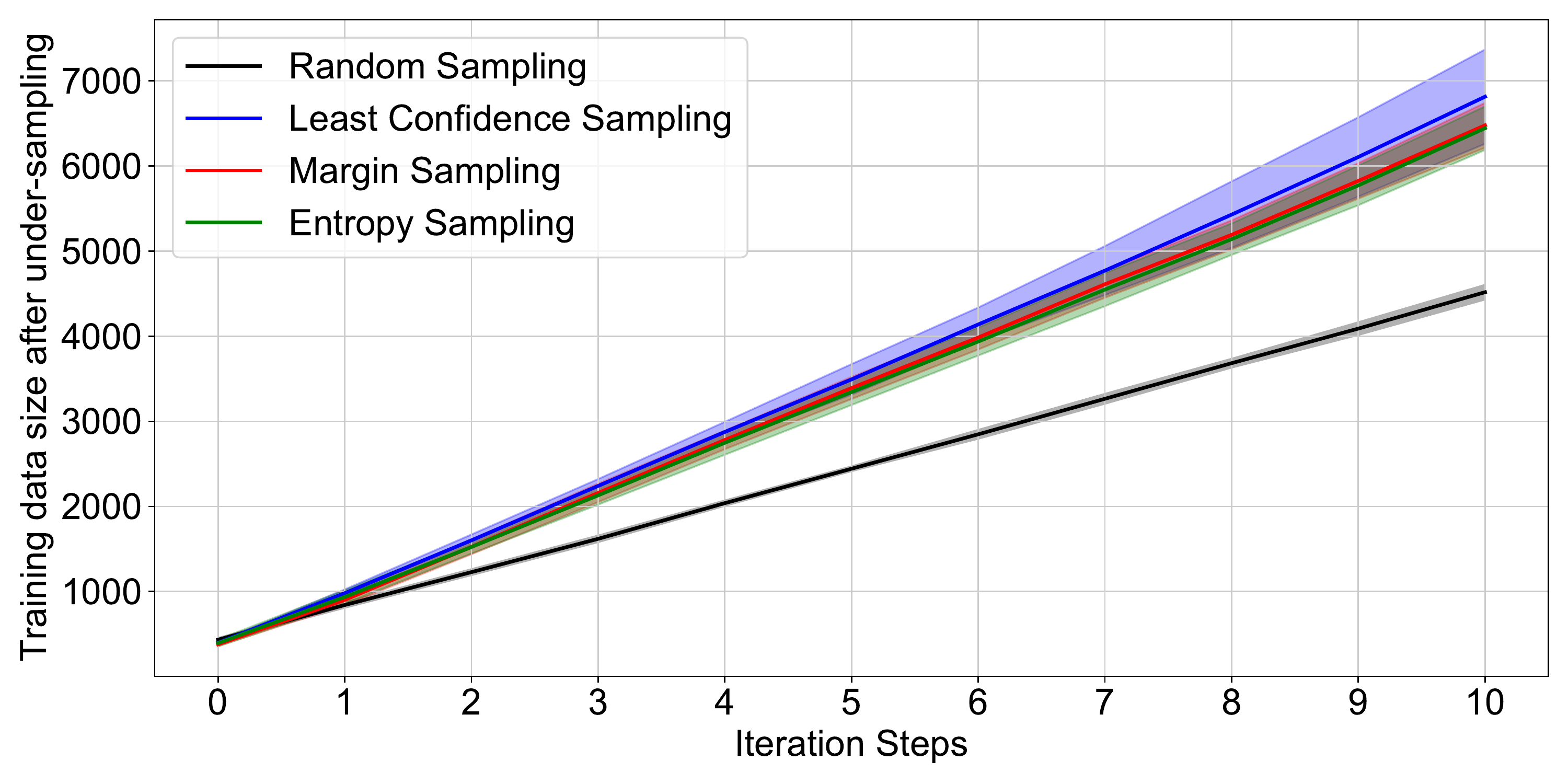}
	\caption{The actual accumulated size of training dataset after under-sampling. The active learner can build up a more balanced training dataset as they can sample towards the decision boundary.}
	\label{fig_al_data}
\end{figure}

Second, a high-dimensional scenario is illustrated. Except for the slack bus (Generator 39 at Bus 10), active and reactive power outputs of all generators are sampled uniformly between the dispatchable limits. Meanwhile, active and reactive power demands of all loads are sampled using normal distributions, which use the base values as the means and admit 50\% standard deviation. In total, we have 57 features. All samples are normalized, among which 80\% samples are allocated for training and 20\% samples for testing. In the active learning, 2000 samples are randomly selected for the initial training phase followed by 2000-sample query iterations. We perform ten iterations and compare all the aforementioned sample strategies, including random (baseline), least-confident, margin, and entropy. We conduct five runs with different random seeds and illustrate the results in Fig. \ref{fig_al}. All three active learning methods have the similar performance, and are all superior to the random sampling. Compared with the random strategy, active learning achieves mostly 5\% accuracy improvement. The actual accumulated size of training dataset after under-sampling is plotted in Fig. \ref{fig_al_data}. In the initial step (Step 0), all strategies randomly select 200 samples, which admit to approximately 400 samples after being under-sampled. Then, active learner can build up a more balanced training dataset as the actual accumulated sizes of training dataset are larger than the random one. This, from another angle, verified than the active learner can sample towards the decision boundary, which could potentially resolve the data imbalance issue.

\section{Conclusions}
This paper proposes the deep active learning method for power system solvability prediction with full AC power flow models. In this problem, sampling over the full power injection space is necessary, which results in a high volume of data to be labeled. To achieve higher labeling and training efficiency, the active learning method is employed, where the most informative instances are selected to be labeled. This method allows us to achieve higher accuracy with much fewer labeled examples. The sampling effectiveness is first visualized in a two-dimensional case. Then, four different sampling strategies are then compared in the high-dimensional solvability prediction. The results indicate that active learning significantly outperforms passive methods and can resolve the the data imbalance issue.

\bibliography{IEEEabrv_zyc,library,Ref_AL_SR}
\bibliographystyle{IEEEtran}

\end{document}